\documentclass[10pt,twocolumn,letterpaper]{article}

\usepackage{wacv}              

\usepackage{graphicx}
\usepackage{amsmath}
\usepackage{amssymb}
\usepackage{bm}
\usepackage{booktabs}
\usepackage{bigstrut}
\usepackage{array}
\usepackage{colortbl}
\usepackage[T1]{fontenc}
\usepackage{wrapfig,lipsum}
\usepackage{booktabs}
\usepackage{bigstrut}
\usepackage{multirow}
\usepackage{placeins}
\usepackage{gensymb}
\usepackage{xcolor}
\usepackage{soul}
\usepackage{makecell}

\makeatletter  
\@namedef{ver@everyshi.sty}{}
\makeatother
\usepackage{tikz}
\usepackage{yfonts}
\usepackage{graphicx}
\usepackage{import}
\usepackage{pifont} 
\usepackage{appendix}
\usepackage[accsupp]{axessibility}  
\usepackage{hhline}
\usepackage{stfloats}
\usepackage{setspace}
\usepackage{bigstrut}
\usepackage{atbegshi,picture}   
\usepackage{mathtools}
\usepackage{pgfplots}
\usepackage{pgfplotstable}
\usepgfplotslibrary{polar}
\usepackage[font={footnotesize}]{caption}
\definecolor{iccvblue}{rgb}{0.21,0.49,0.74}
\usepackage[pagebackref,breaklinks,colorlinks,allcolors=iccvblue]{hyperref}

\usepackage[capitalize]{cleveref}
\crefname{section}{Sec.}{Secs.}
\Crefname{section}{Section}{Sections}
\Crefname{table}{Table}{Tables}
\crefname{table}{Tab.}{Tabs.}

\definecolor{darkgreen}{RGB}{0,127,0}
\definecolor{darkred}{RGB}{200,0,0}

{\end{list}}

\def\wacvPaperID{3083} 
\def\confName{WACV}
\def\confYear{2026}

\title{WALDO: Where Unseen Model-based 6D Pose Estimation Meets Occlusion}

\author{$\textit{Sajjad Pakdamansavoji}^*$, \textit{Yintao Ma},  \textit{Amir Rasouli},  \textit{Tongtong Cao}\\
\textit{Huawei Technologies Canada}\\
{\tt\small $\textit{sj.pakdaman.edu@gmail.com}^*$}}

\begin{document}
\maketitle
\begin{abstract}

Accurate 6D object pose estimation is vital for robotics, augmented reality, and scene understanding. For seen objects, high accuracy is often attainable via per-object fine-tuning but generalizing to unseen objects remains a challenge. To address this problem, past arts assume access to CAD models at test time and typically follow a multi-stage pipeline to estimate poses: detect and segment the object, propose an initial pose, and then refine it. Under occlusion, however, the early-stage of such pipelines are prone to errors, which can propagate through the sequential processing, and consequently degrade the performance. To remedy this shortcoming, we  propose four novel extensions to model-based 6D pose estimation methods: (i) a dynamic non-uniform dense sampling strategy that focuses computation on visible regions, reducing occlusion-induced errors; (ii) a multi-hypothesis inference  mechanism that retains several confidence-ranked pose candidates, mitigating brittle single-path failures; (iii) iterative refinement to progressively improve pose accuracy; and (iv) series of occlusion-focused training augmentations that strengthen robustness and generalization. Furthermore, we propose a new weighted by visibility metric for evaluation under occlusion to minimize the bias in the existing protocols. Via extensive empirical evaluations, we show that our proposed approach achieves more than \textbf{5\%} improvement in accuracy on ICBIN and more than \textbf{2\%} on BOP dataset benchmarks, while achieving $\approx 3\times$ faster inference.

\end{abstract}    

\section{Introduction}
\label{sec:intro}

Accurate 6D object pose estimation is a fundamental problem in various domains, such as robotics, augmented reality, and scene understanding. 
Over the past decade a wide range of methods have been proposed, including model-free methods that only rely on images or videos of the objects \cite{hodan2024bop} or model-based methods that resort to using CAD models for more generalability to unseen objects \cite{lin2024sam}. These innovations have been supported by a wide range of techniques, including  synthetic data generation, including large-scale 3D asset repositories \cite{chang2015shapenet, downs2022google, deitke2023objaverse}, photorealistic rendering pipelines \cite{to2018ndds, denninger2019blenderproc}, and neural architectural innovations \cite{qin2022geometric, zhao2021point}. These methods enabled large-scale pretraining, leading to better generalization to out-of-distribution (OOD) scenarios as well as narrowing the sim-to-real gap.



\begin{figure}
    \centering
    \includegraphics[width=1\linewidth]{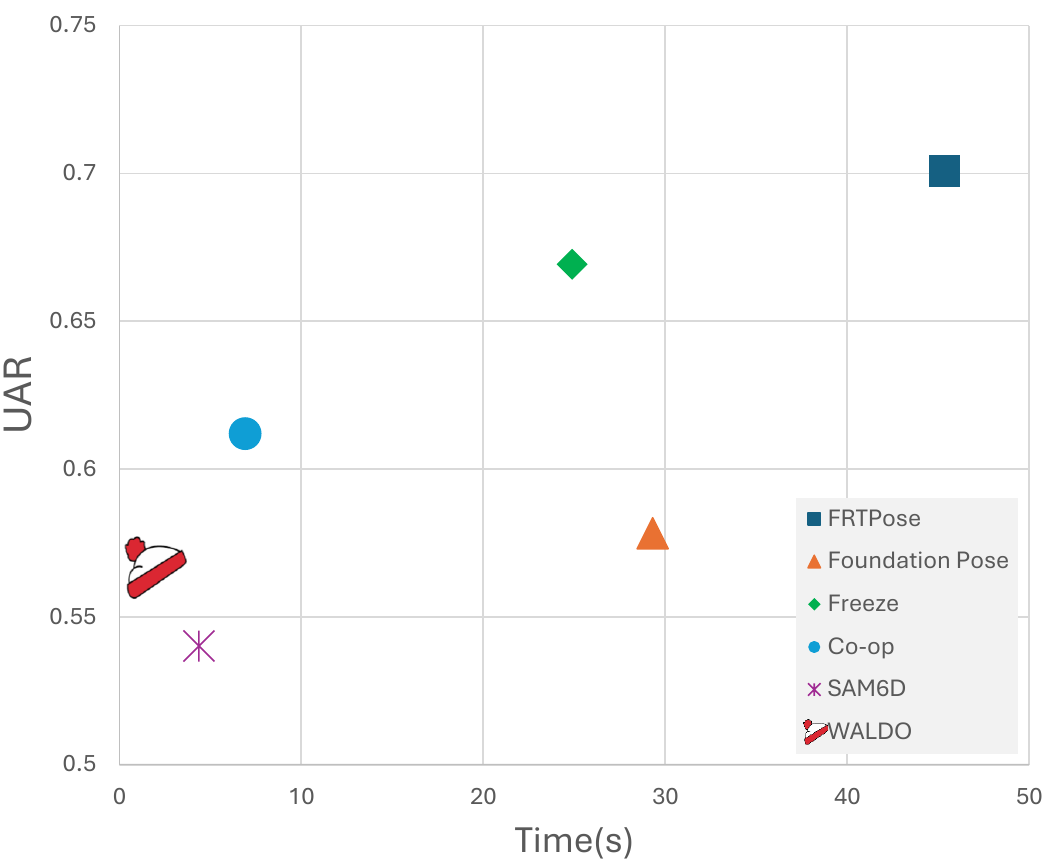}
    \caption{Scatter plot comparing unbiased average recall (UAR) versus inference time for multiple methods on the BOP-Core* dataset. WALDO performs better compared to past arts on both metrics.}
    \label{fig:UAR_BOP_Core_all} \vspace{-0.4cm}
\end{figure}

Despite the substantial progress, accurate 6D pose estimation under practical conditions involving clutter and occlusion remains an open problem, in particular for unseen objects. In the unseen object setting state-of-the-art methods assume a CAD model at the test time and typically follow a detect-segment-pose-refine pipeline. However, under occlusion, errors introduced in earlier stage would potentially accumulate and degrade pose estimation accuracy.

To remedy this problem, several solutions have been proposed, including domain randomization \cite{legrand2024domain}, additional modalities incorporation \cite{castro2023posematcher}, local-feature extraction and search \cite{wen2024foundationpose}, and leveraging large foundation models \cite{caraffa2024freeze}. However, these approaches typically address occlusion only indirectly, are computationally expensive, and often depend on strong assumptions about the environment, limiting their practical use in previously unexplored scenes.

Besides the limitations of the existing approaches, evaluation protocols are not adequate for assessment under occlusion. The existing benchmarks \cite{hodan2024bop} lack systematic scene variation and their metrics do not account for the occlusion distribution in the data. Hence, the reported results are biased toward high-visibility cases, inflating reported performance and obscuring the true impact of occlusion.

In this work, we address the problem of 6D pose estimation under occlusion by coupling inference-time strategies that focus on visible regions of the object with training-time augmentations that simulate partial visibility. To navigate pose uncertainty, we maintain multiple initial hypotheses and iteratively refine them, yielding higher accuracy. Finally, we propose an occlusion-aware variant of standard evaluation metrics that provides unbiased performance estimates across visibility levels.

In summary, the contributions of our work are:

\begin{itemize}
    \item We address occlusion by estimating occlusion probabilities while forming the initial pose hypothesis and then use them to prioritize visible regions during refinement.
    
    \item To manage pose uncertainty, we generate multiple initial hypotheses, refine them iteratively, and select the final pose based on its post-refinement confidence score.

    \item We propose training-time augmentations specifically designed to simulate occlusion and improve robustness to partial visibility.
    
    \item We propose occlusion-aware evaluation metrics for 2D detection, segmentation, and 6D pose estimation that provide unbiased performance across visibility levels.

\end{itemize}

\section{Related Work}
\label{sec:related_work}
\textbf{6D Pose Estimation.} models can be categorized based on wether the objects are \emph{seen} during training or are \emph{unseen}. For \emph{seen} objects, early systems \cite{rad2017bb8,kehl2017ssd} were trained per-instance to learn extracting stable 2D--3D cues, such as keypoints \cite{peng2019pvnet}, view/rotation labels \cite{xiang2018posecnn}, dense surface codes \cite{haugaard2022surfemb}, and geometry-guided predictors \cite{wang2021gdr,su2022zebrapose,yang2023scflow}. 
using exact CAD models and per-object fine-tuning, these frameworks can reach high accuracy even under significant occlusion because correspondence supervision tightly binds image evidence to the specific mesh \cite{hodan2020epos}.

Generalizing to \emph{unseen} instances at test time is notably harder, especially under occlusion because overfitting to specific shapes or contexts does not transfer. \emph{Model-free} methods attempt to remedy this problem by building an object-specific reference bank using only a few views of the objects and estimate their pose at run-time by local feature matching and refinement \cite{sun2022onepose,liu2022gen6d}. However, using an incomplete reference model can lead to failure under occlusion due to sparse observation and weak matching results.


 To remedy this issue, \emph{model-based} resort to using object CAD models. Some use a single-pass 6D pose estimation at inference, using direct regression/classification \cite{tekin2018real} or implicit pose-code retrieval \cite{sundermeyer2018implicit}. This approach, however, comes at a cost. Due to the lack of a self-correction mechanism, error from imperfect detection or occluder confusion cannot be recovered, therefore accuracy degrades as occlusion increases.

Two-stage methods solve this problem by,  first deriving an initial pose and then refining it. For instance, the model in \cite{labbemegapose} predicts the correspondences between 
 keypoints or dense object coordinates and estimates an initial pose via solving a Perspective-n-Point (PnP) \cite{PnP} problem. Other methods extend this by adding a  refinement module to align a rendered view to the image (render-and-compare) \cite{li2018deepim} or fuse depth with learned features \cite{wang2019densefusion} to reduce residual error. Such architectures allow self-correction under occlusion where the refiner can repair mislabeled evidence using the CAD prior. However, because the pipeline is multi-stage, weak initial hypothesis can still propagate and limit final accuracy unless hypothesis diversity and scoring are carefully designed \cite{labbe2020cosypose}. In this work, we adopt a model-based approach and use a novel sampling method to mitigate the error propagation problem. 


\begin{figure*}[t]
    \centering
    \includegraphics[width=1\linewidth]{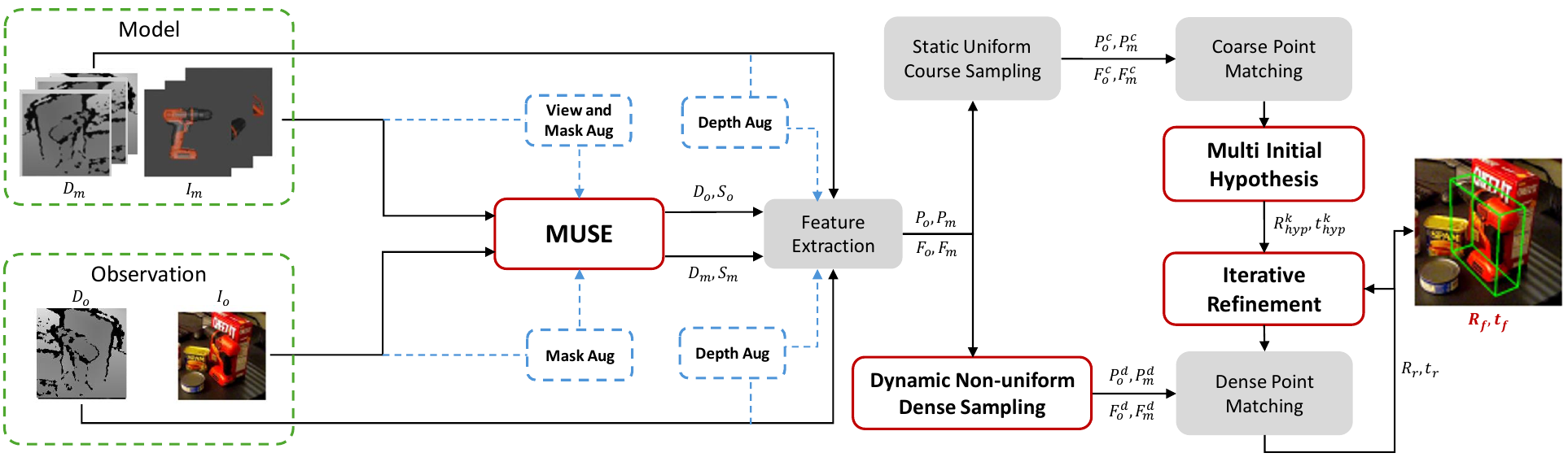}
    \caption{Overview of our 6D pose estimation framework. The observed RGB–D image and the rendered RGB–D template are first sent to MUSE to obtain bounding boxes and masks. Using these, the inputs are processed and passed through a feature extractor to compute embeddings and their corresponding point clouds. A uniform coarse sample of the point clouds goes to a coarse point-matching module, which produces multiple initial pose hypotheses and per-point occlusion probabilities. Guided by these probabilities, we perform dynamic non-uniform dense sampling and iteratively refine the final pose.}
    \label{fig:framework}
    \vspace{-1.3em}
\end{figure*}

\textbf{Occlusion Handling} is a core challenge in 6D pose estimation. A range of strategies have been explored to address this problem. Domain randomization methods alter textures, lighting, and backgrounds \cite{legrand2024domain, ikeda2022sim2real, chen2023texpose} or generate novel views with Neural Radiant Fields \cite{mildenhall2021nerf}, encouraging models to learn appearance-invariant features. However, these augmentations primarily improve generalization and have only indirect effects on robustness to occlusion. Other methods use additional modalities such as depth to align objects based on 3D shape cues \cite{castro2023posematcher}, but these signals are not always available in practice.

Some works explicitly model occlusion. For example, \cite{lin2024sam} identifies regions of the object as background and excludes them from matching. Other approaches \cite{sun2022onepose, wang2024stable} design architectures for local feature extraction and matching, enabling reliable pose recovery using only visible parts. A different line of work \cite{wen2024foundationpose, labbemegapose} employs local search through render-and-compare strategies to handle partial observations, though at the cost of high computation \cite{iwase2021repose}.

More recently, large foundation models have been applied to occlusion-aware pose estimation. These methods \cite{caraffa2024freeze, caraffa2025accurate, kai2024vfm, pos3R} extract robust representations from partially visible inputs and generalize well, but they remain limited under severe occlusion and often require slow inference.

\textbf{Data and Evaluation Protocols.}
Modern 6D pose data is based on a small set of canonical collection of scenes with objects placed in various poses, many of which are included in the BOP benchmark \cite{hodan2024bop}. Together they span key axes of variability, including texture, symmetry, modality, clutter, and occlusion. Despite efforts made to capture occlusion, the datasets are still highly skewed towards highly visible object views, consequently biasing training. On evaluation side, commonly used metrics do not account for such a bias. Metrics, such as Visible Surface Discrepancy (VSD),  Maximum Symmetry-Aware Surface Distance (MSSD), and Maximum Symmetry-Aware Projection Distance (MSPD) focus on measuring the discrepancy and symmetry of visible parts of objects, failing to account for the degree of occlusion, mirroring the inherent biases in the datasets. We address this issue by proposing a weighted averaging method to minimize data bias for more balanced ranking of models.

\section{Approach}
Our proposed framework, WALDO, consists of three parts: \textbf{3D-aware object detection} which locates and masks the target object; \textbf{Feature extraction}, which draws local features from the masked views and lifts sampled features to 3D; \textbf{Pose estimation}, which generates initial pose hypotheses from a coarse point cloud and refines them with the dense point cloud to produce the final 6D pose. An overview of our framework is given in Fig. \ref{fig:framework}. The details of these modules are given below.
\subsection{3D-aware Object Detection}
\label{subsec:detection}
The first step in recovering the 6D pose is to detect the target object in the observations. Following MUSE~\cite{muse}, we use Grounding~DINO~\cite{liu2023grounding} on the observed image \(\mathcal{I}_o\) to propose bounding boxes, then apply SAM2’s box prompt~\cite{lin2024sam} to obtain segmentation masks. The foreground region selected by the mask is encoded with DINOv2-L~\cite{oquab2023dinov2} to derive multi-level embeddings that capture both class-level and patch-level cues. The same encoder processes the rendered model templates \(\{\mathcal{I}_m^{v}\}_{v=1}^{V}\). Template embeddings are weighted with an Integrated von~Mises–Fisher (I\mbox{-}vMF) scheme~\cite{banerjee05a} to favor plausible viewpoints. Next, we compute the 3D-aware multi-embedding similarity between the observation and the templates features to correctly match them. Lastly, we return bounding boxes and segmentation masks for both observation \((\mathcal{B}_o,\mathcal{S}_o)\) and model \(\{(\mathcal{B}_m^{v}, \mathcal{S}_m^{v})\}_{v=1}^{V}\).

\subsection{Feature Extraction}
\label{subsec:feature_extraction}

Given \((\mathcal{B}_o,\mathcal{S}_o)\), we process $\mathcal{I}_{o}$ by masking out the background, cropping the region of interest, and resizing the crop to a fixed resolution (e.g., $224\times224$). The resulting image is fed into a Visual Transformer (ViT) backbone \cite{dosovitskiy2020image} to extract per-patch embeddings at multiple depths (e.g., after the 3rd, 6th, 9th, and 12th blocks). The resulting patch embeddings are then transformed via a fully-connected layer, reshaped, and bilinearly interpolated to form a feature map \(
    \mathcal{F}_o \in \mathbb{R}^{N_o\times C} \),
where $N_o$ is the number of pixels in $\mathcal{S}_o$ (or points) and $C$ is the feature dimension.

The same process is repeated for each $\mathcal{I}_m^v$ and \((\mathcal{B}_m^v,\mathcal{S}_m^v)\), yielding per-template feature maps $\mathcal{F}_m^v \in \mathbb{R}^{N_m^v\times C}$ where $N_m^v$ is the number of pixels in $\mathcal{S}_m^v$. Per-template feature maps are then aggregated to form $\mathcal{F}_m \in \mathbb{R}^{N_m\times C}$ where $N_m=\sum_{v=1}^V N_m^v$. To model segmentation imperfections  and partial visibility, we append learnable background ${f}^{bg}_{o}\in\mathbb{R}^{C}$ and occlusion ${f}^{oc}_{m}\in\mathbb{R}^{C}$ embeddings to $\mathcal{F}_o$ and $\mathcal{F}_m$, respectively.


Given the observed depth $\mathcal{D}_{o}$ and $\mathcal{S}_o$ we build a partial point cloud  $\mathcal{P}_o$. Similarly with model's rendered depth $\{ \mathcal{D}_m^v \}_{v=1}^V$ and $\{ \mathcal{S}_m^v \}_{v=1}^V$ a full point cloud of the model $\mathcal{P}_m$ is attainable. Using these point clouds directly, imposes large computational overhead; therefore, we use a down sampled version using one of the following options.

\textbf{Static Uniform Sampling}: One way to achieve this is to uniformly sample from the original point cloud. In practice, this is done by iterative farthest point sampling \cite{qi2017pointnet++} to build a set that have the largest minimum pairwise distance, thereby approximating a uniform sample.

\textbf{Dynamic Non-uniform Sampling}: Given per-point probabilities for a point cloud, another way is to build a Walker–Vose alias table \cite{walker1977efficient} and apply the Gumbel-Top-k trick \cite{kool2019stochastic} to sample $k$ points without replacement, thereby approximating a sample aligned to the given probabilities.

\begin{figure}
    \centering
    \includegraphics[width=1\linewidth]{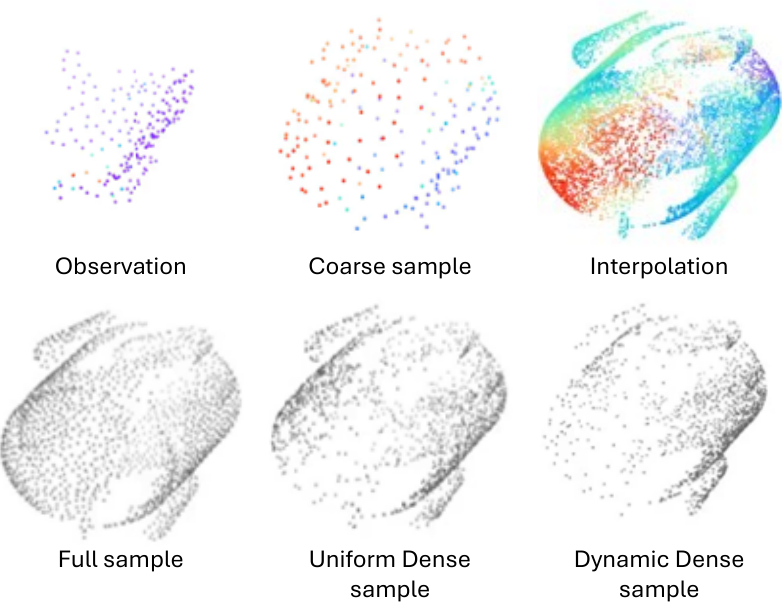}
    \caption{\textbf{(Top)} Estimated occlusion/background probabilities visualized by color (red = higher probability). Comparing the observation with the coarse sample indicates accurate probability estimates; interpolation extends these estimates to the full point cloud. \textbf{(Bottom)} Probability-guided dynamic non-uniform sampling allocates more points to visible regions of the object compared to conventional static uniform sampling.}
    \label{fig:combined}\vspace{-0.4cm}
\end{figure}
\subsection{6D Pose Estimation}
\label{subsec:pose_estimation_module}

We solve pose estimation via \emph{3D–3D} correspondences (point matching) between the observation and model point clouds. We first use a coarse point-cloud sample to generate multiple initial hypotheses, and then iteratively refine them to obtain an accurate pose. The details of this process is given below.

\subsubsection{3D-3D Correspondence}
We estimate a \emph{3D-3D} correspondence matrix between the observation and the model from their corresponding course/dense point clouds, and then obtain the final pose by applying SVD to this matrix. Concretely, we compute an attention matrix $\mathcal{A}$ (Eq. \ref{eq:assignment}), then normalize it with a temperature-scaled ($\tau$) softmax applied along both rows and columns to produce $\tilde{\mathcal{A}}$ (Eq. \ref{eq:soft_assignment}). The resulting $\tilde{\mathcal{A}}$ encodes correspondence probabilities between observation and model points, and maps non-overlapping regions to background on the observation side and to occluded on the model side.

\begin{equation}
    \mathcal{A} = \begin{bmatrix}{f}^{bg}_{o}, \mathcal{F}_o\end{bmatrix} \times \begin{bmatrix}{f}^{oc}_{m}, \mathcal{F}_m\end{bmatrix}^T \in \mathbb{R}^{(N_o+1)\times (N_m+1)}
    \label{eq:assignment}
\end{equation}

\begin{equation}
    \tilde{\mathcal{A}} = \operatorname{Softmax}_{\text{row}}\Bigl(\frac{\mathcal{A}}{\tau}\Bigr) \cdot \operatorname{Softmax}_{\text{col}}\Bigl(\frac{\mathcal{A}}{\tau}\Bigr).
    \label{eq:soft_assignment}
\end{equation}


\subsubsection{Coarse Point Matching}  
We sample coarse point sets uniformly from the observation \( \mathcal{P}_o^c \in \mathbb{R}^{N_o^c \times 3} \) and the object's model \( \mathcal{P}_m^c \in \mathbb{R}^{N_m^c \times 3} \). These course sets, along with their corresponding features $\mathcal{F}_o^c$ and $\mathcal{F}_m^c$ are further processed by a stack of $T^c$ Geometric Transformers \cite{qin2022geometric} that enhance the feature representations and facilitate intra-set and inter-set interactions. The coarse correspondence matrix $\tilde{\mathcal{A}}^c$ is then derived from these features.  We set $N_b^c = N_o^c = 196$ and $T^c=3$.

\subsubsection{Multi Initial Pose Hypothesis}
The course assignment matrix $\tilde{\mathcal{A}}^c$ is used to draw multiple samples of correspondences and generate multiple coarse pose hypotheses $ (\mathcal{R}_{\text{hyp}}^{k} , t_{\text{hyp}}^k )$ via SVD. Each hypothesis is given a confidence score $s_{\text{hyp}}^k$ (Eq. \ref{eq:pose_confidence}) based on its average intra-set distance to the model point cloud. The top $K$ hypothesis selected based on confidence are passed to the Dense Point Matching module.

\begin{equation}
\label{eq:pose_confidence}
    s_{\text{hyp}} = \frac{N_o^c}{\sum_{\bm{p}_o^c \in \mathcal{P}_o^c} \min_{\bm{p}_m^c \in \mathcal{P}_m^c} \left\| \mathcal{R}_{\text{hyp}}^k (\bm{p}_m^c - t_{\text{hyp}}^k) - \bm{p}_o^c \right\|_2},
\end{equation}

\subsubsection{Dynamic Non-uniform Dense Sampling}
The first column and row of $\tilde{\mathcal{A}}^{c}$ provide occlusion and background probabilities for the coarse samples, denoted $Pr_{\mathrm{oc}}(\mathcal{P}_m^{c})$ and $Pr_{\mathrm{bg}}(\mathcal{P}_o^{c})$. To obtain probabilities for the original point clouds, $Pr_{\mathrm{oc}}(\mathcal{P}_m)$ and $Pr_{\mathrm{bg}}(\mathcal{P}_o)$, we apply weighted linear interpolation from the coarse samples, using weights inversely proportional to the 3D Euclidean distance between points and normalized to sum to one. We further use these probabilities to draw a dynamic non-uniform dense sample of the point clouds $ \mathcal{P}_o^d \in \mathbb{R}^{N_o^d \times 3}$ and $\mathcal{P}_m^d \in \mathbb{R}^{N_m^d \times 3}$. Figure \ref{fig:combined} visualizes this process.

\begin{equation}
\Pr_{p}(Q[i]) \;=\; \sum_{j=1}^{\tilde N} 
\frac{\|Q[i]-Q^c[j]\|_2}{\sum_{k=1}^{\tilde N}\|Q[i]-Q^c[k]\|_2}\,\Pr_{p}(Q^c[j]),
\end{equation}
\[
\text{where } p \in \{oc, bg\}, \quad Q \in \{\mathcal{P}_m, \mathcal{P}_o\}, \quad Q^c \in \{\mathcal{P}_m^c, \mathcal{P}_o^c\}.
\]

\subsubsection{Dense Point Matching}  
Prior to dense matching, we encode positional information by applying a multi-scale Set Abstraction module \cite{qi2017pointnet++} to the points, where $\mathcal{P}_o^d$ is transformed by the coarse pose $(\mathcal{R}_{hyp}, t_{hyp})$. These positional encodings are then added to the dense features $\mathcal{F}_o^d$ and $\mathcal{F}_m^d$. Next, the extended features are  passed through a series of $T^d$ Sparse-to-Dense Point Transformers \cite{lin2024sam}. From the assignment matrix $\tilde{\mathcal{A}}^d$, dense correspondences are obtained and the refined pose $(\mathcal{R}_r, t_r)$ is computed via weighted SVD. We set $N_b^d = N_o^d = 2048$ and $T^d=3$.

\subsubsection{Iterative Pose Refinement}  
To improve the accuracy of the estimated pose, we adopt an iterative refinement strategy. Specifically, the refined pose $(\mathcal{R}_r, t_r)$ obtained from the dense point matching stage is reused as a new hypothesis $(\mathcal{R}_{hyp}, t_{hyp})$ for the next iteration. The object's dense point cloud $\mathcal{P}_o^d$ is transformed again using this updated hypothesis, and the dense point matching module is repeated. This process is applied for $N$ refinement steps, progressively reducing pose errors. After $N$ iterations, the final pose estimate is denoted as $(\mathcal{R}_f, t_f)$.
\subsubsection{Occlusion-focused Augmentation}
\label{subsec:augmentations}

\begin{figure}[t]
  \centering
  \begin{subfigure}{\linewidth}
    \includegraphics[width=\linewidth]{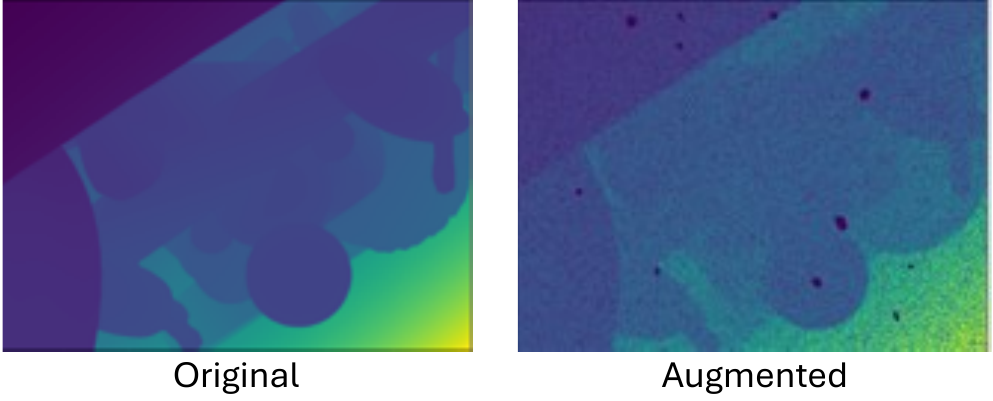}
    \label{fig:depth_aug_example}
  \end{subfigure}
  \vspace{-4mm}

  \begin{subfigure}{\linewidth}
    \includegraphics[width=\linewidth]{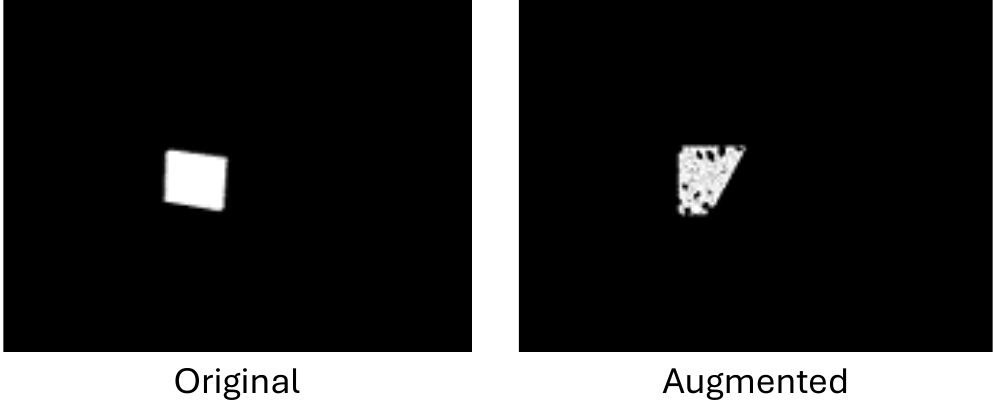}
    \label{fig:mask_aug_example}
  \end{subfigure}
  \vspace{-4mm}

  \begin{subfigure}{\linewidth}
    \includegraphics[width=\linewidth]{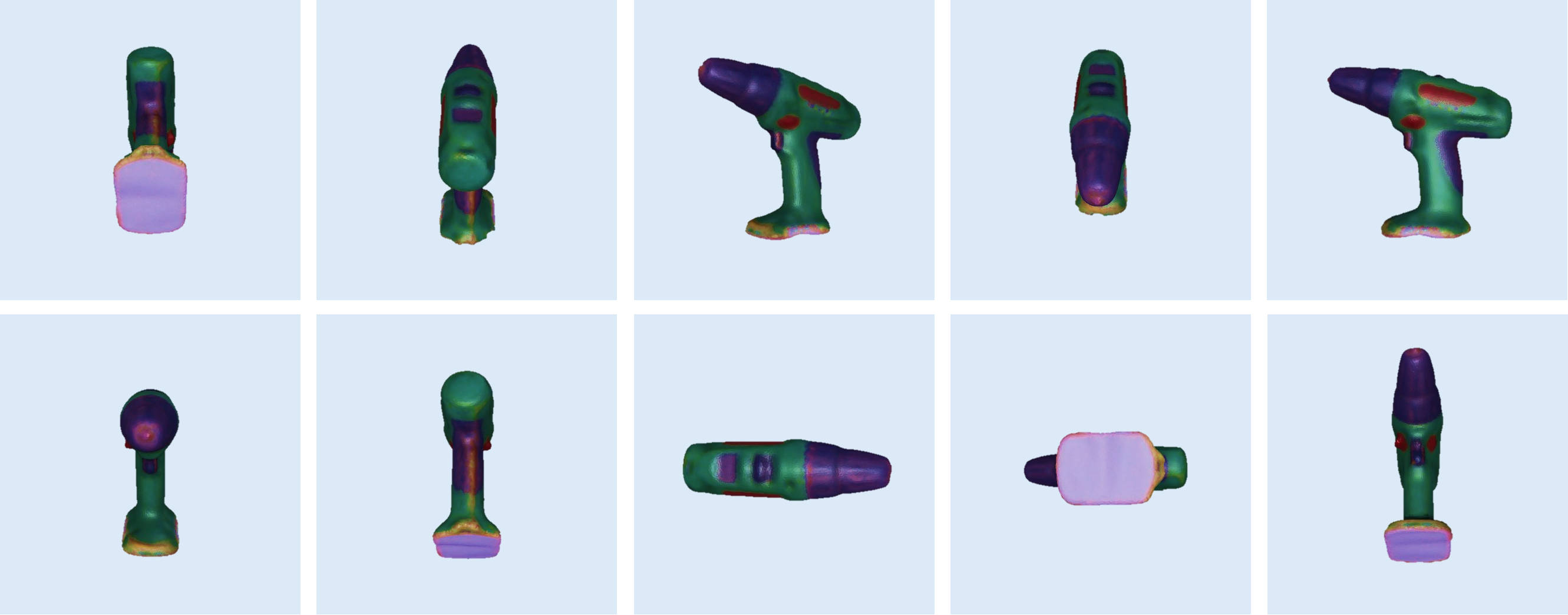}
    \label{fig:templates_view}
  \end{subfigure}
  \vspace{-6mm}
  \caption{Qualitative examples of our proposed \textbf{(Top)} depth augmentation, \textbf{(Middle)} mask augmentation, and \textbf{(Bottom)} object templates rendered from different viewpoints.}
  \label{fig:all_aug_templates}
\end{figure}

To generalize to unseen objects, we train on a large-scale synthetic dataset \cite{labbemegapose}. To narrow the sim-to-real gap, we apply standard photometric (e.g., color jitter, noise, blur) and geometric (e.g., scaling, rotation, perspective) augmentations. To explicitly simulate occlusion, we further propose the following occlusion-focused augmentations. Qualitative examples are given in Fig. \ref{fig:all_aug_templates}.

\begin{itemize}
\item \textbf{Depth Augmentation:} Real depth sensors often produce noise and missing values, especially under occlusion and clutter. To mimic this, we follow \cite{wen2024foundationpose} and apply blurring, Gaussian noise, and random depth dropouts, either pixel-wise or in elliptical regions.

\item \textbf{Mask Augmentation.} Predicted masks are rarely perfect, often include background or missing object parts. These issues are further amplified by occlusion. We simulate these errors by dilating masks, replacing them with bounding boxes, splitting them with straight lines, and randomly dropping internal regions.

\item \textbf{View Augmentation.} Inspired by what \cite{lin2024sam} does at inference, we render 42 model templates (24 pairs with opposing views). During training, 1–2 views are randomly sampled from each pair, producing partial occlusion in the constructed model point cloud.
\end{itemize}

\subsubsection{Training Objective}
\label{subsec:training_obj}

\paragraph{Correspondence Supervision.}
To supervise \emph{3D-3D} correspondences between the observation and the object point clouds, we adopt an InfoNCE objective \cite{oord2018representation} implemented as bidirectional cross-entropy over the correspondence matrix. Let $\mathcal{A}$ be the matrix from Eq.~\ref{eq:assignment}. The loss is defined as
\begin{equation}
    \mathcal{L} = \texttt{CE}\Bigl(\mathcal{A}[1:, :], \hat{\mathcal{Y}}_o\Bigr) + \texttt{CE}\Bigl((\mathcal{A}[:, 1:])^T, \hat{\mathcal{Y}}_m\Bigr),
    \label{eq:loss}
\end{equation}
where $\texttt{CE}(\cdot,\cdot)$ is cross-entropy, $\hat{\mathcal{Y}}_o\in\mathbb{R}^{N_o}$ are observation-to-model labels (one per observed point), and $\hat{\mathcal{Y}}_m\in\mathbb{R}^{N_m}$ are model-to-observation labels (one per template point). The first term classifies each observed point against all model indices (including the background token), while the second term  classifies each model point against all observed indices (including the occlusion token).

\paragraph{Ground-truth labels.}
Ground-truth labels are obtained from the known pose $(\hat{\mathcal{R}}, \hat{t}\,)$ via nearest-neighbor assignment. For each observation point $\bm{p}_o\in \mathcal{P}_o$,
\begin{equation}
    \hat{y}_o = \begin{cases}
      0,   & \text{if } d_{m^*} \ge \delta_{dis}, \\
      m^*, & \text{if } d_{m^*} < \delta_{dis},
    \end{cases}
\end{equation}
with
\begin{equation}
    m^* = \operatorname*{argmin}_{m=1,\dots,N_m} \left\| \hat{\mathcal{R}}(\bm{p}_o - \hat{t}\,) - \bm{p}_{m} \right\|_2,
\end{equation}
and
\begin{equation}
    d_{m^*} = \left\| \hat{\mathcal{R}}(\bm{p}_o - \hat{t}) - \bm{p}_{m^*} \right\|_2.
\end{equation}
Here, $\delta_{dis}$ is a fixed threshold set to $0.15$ empirically and $\hat{y}_o = 0$ denotes background. An analogous construction yields $\hat{y}_m$ for object's model points where $\hat{y}_m = 0$ denotes occluded.

\paragraph{Final Multi-block objective.}
Finally, we aggregate losses across transformer blocks in both the coarse and dense matching stages. The final loss is given in Eq. \ref{eq:final_loss} where $\mathcal{L}_l^c$ and $\mathcal{L}_l^d$ are the losses at the $l$-th blocks of the course and dense stages respectively.
\begin{equation}
\label{eq:final_loss}
    \hat{\mathcal{L}}= \min \sum_{l=1}^{T^c} \mathcal{L}_l^c \;+\; \sum_{l=1}^{T^d} \mathcal{L}_l^d,
\end{equation}

\section{Experiments}

\textbf{Dataset.}We evaluate our proposed method on five common datasets from BOP benchmark: LMO \cite{brachmann2014learning}, YCB-V \cite{xiang2018posecnn}, T-LESS \cite{hodan2017t}, IC-BIN \cite{doumanoglou2016recovering}, and TUD-L \cite{hodan2024bop}--together denoted as BOP-Core*. These datasets include dense clutter, multiple instances, static and dynamic scenes, and tabletop or robotic manipulation, with objects that are textureless, reflectant, symmetric, and of varied sizes.

\begin{figure}[t]
    \includegraphics[width=\columnwidth]{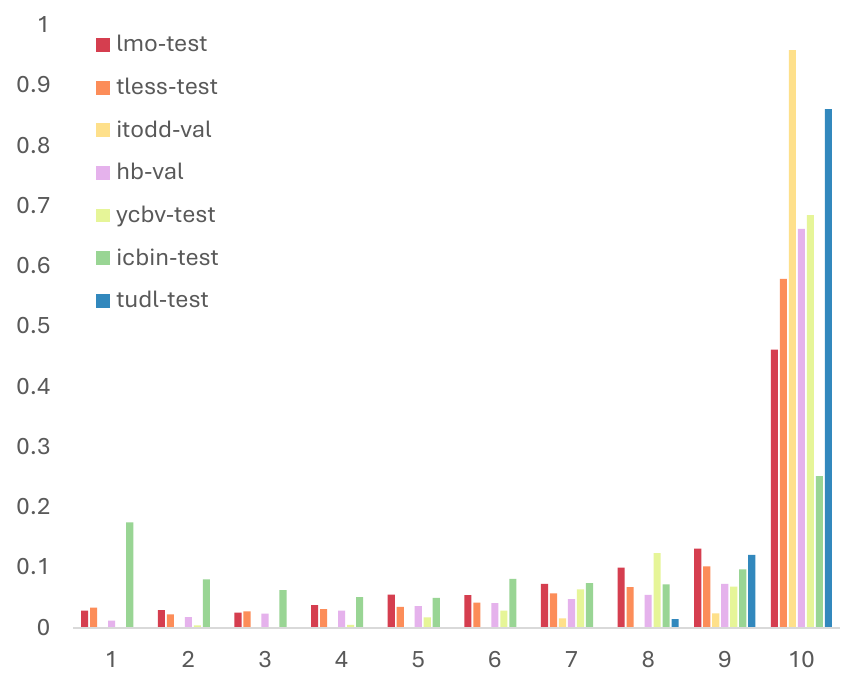}
    \caption{Ratio of instances in the BOP-Core datasets divided into 10 deciles based on their visibility fraction.}
    \label{fig:BOP_visibility}
\end{figure}

\noindent\textbf{Metrics.} 6D pose estimation is commonly evaluated using the BOP Average Recall (AR), computed over VSD, MSSD, and MSPD \cite{hodan2018bop}. As shown in Fig. \ref{fig:BOP_visibility}, many datasets skew toward non-challenging highly visible instances (visibility fraction $>90\%$), which inflates reported performance. To remedy this, we propose Unbiased Average Recall (UAR): we group test instances into ten visibility deciles (set empirically), compute recall within each decile for VSD, MSSD, and MSPD, and then average these decile recalls across deciles and metrics. Therefore, UAR acts as a single score that reflects performance across all occlusion levels rather than being biased to easy, highly visible cases.

Detection and segmentation performance is often reported by COCO Average Precision (AP) and Average Recall (AR) \cite{lin2014microsoft}. COCO already stratifies the results according to object size (small, medium, large); we extend this idea to visibility by computing AP and AR within each visibility decile and take an average across them. These decile-based AP/AR metrics, denoted as mAPD and mARD, provide a fairer assessment under occlusion by ensuring that low-visibility instances contribute equally to the score.

\noindent\textbf{Models.} We primarily compare against our baseline, SAM6D \cite{lin2024sam}. For completeness, we also report state-of-the-art methods from the BOP benchmark leaderboard: for detection, CNOS \cite{nguyen2023cnos}, Fast-SAM \cite{ravi2024sam}, and NIDS; and for pose estimation, FRTPose, FoundationPose \cite{wen2024foundationpose}, FreeZe \cite{caraffa2024freeze}, and Co-op.

\subsection{Detection and segmentation}\label{sec:detection}
We compare detection and segmentation performance on BOP-Core* against commonly used open-vocabulary models and, in particular, our base model SAM6D (see Table~\ref{tab:detection_precision}). Our system achieves \(mAPD=0.218\) and \(mARD = 0.444\), improving over SAM6D by \(1.6\%\) and \(1.1\%\), respectively. Notably, conventional \(AP\) and \(AR\) overestimate performance by overweighting highly visible objects: for WALDO, \(AP\) exceeds \(mAPD\) by \(+0.29\) (\(\approx 133\%\)) and \(AR\) is higher than \(mARD\) by \(+0.15\) (\(\approx 34\%\)), a pattern consistent across all models.

\begin{table}
    \centering
    \caption{Detection and segmentation on BOP-Core*. We report the standard \(AP\) and \(AR\) alongside the proposed \(mAPD\) and \(mARD\).}
    \begin{tabular}{l | c c | c c}
      \toprule
       Model & AP & mAPD & AR & mARD\\
      \midrule
       CNOS     & 0.425 & 0.169 & 0.519 &  0.374\\
       Fast-SAM & 0.456 & 0.216 & 0.528 &  0.384\\
       NIDS     & 0.463 & 0.214 & 0.555 &  0.421\\
       \midrule
       SAM6D    & 0.491 & 0.202 & 0.576 &  0.433\\
       \textbf{WALDO}     & \textbf{0.508} & \textbf{0.218} & \textbf{0.594} &  \textbf{0.444}\\
      \bottomrule
    \end{tabular}\vspace{-0.4cm}
    \label{tab:detection_precision}
\end{table}

\subsection{6D pose estimation}\label{sec:pose}
\begin{figure*}
    \centering
    \includegraphics[width=1\linewidth]{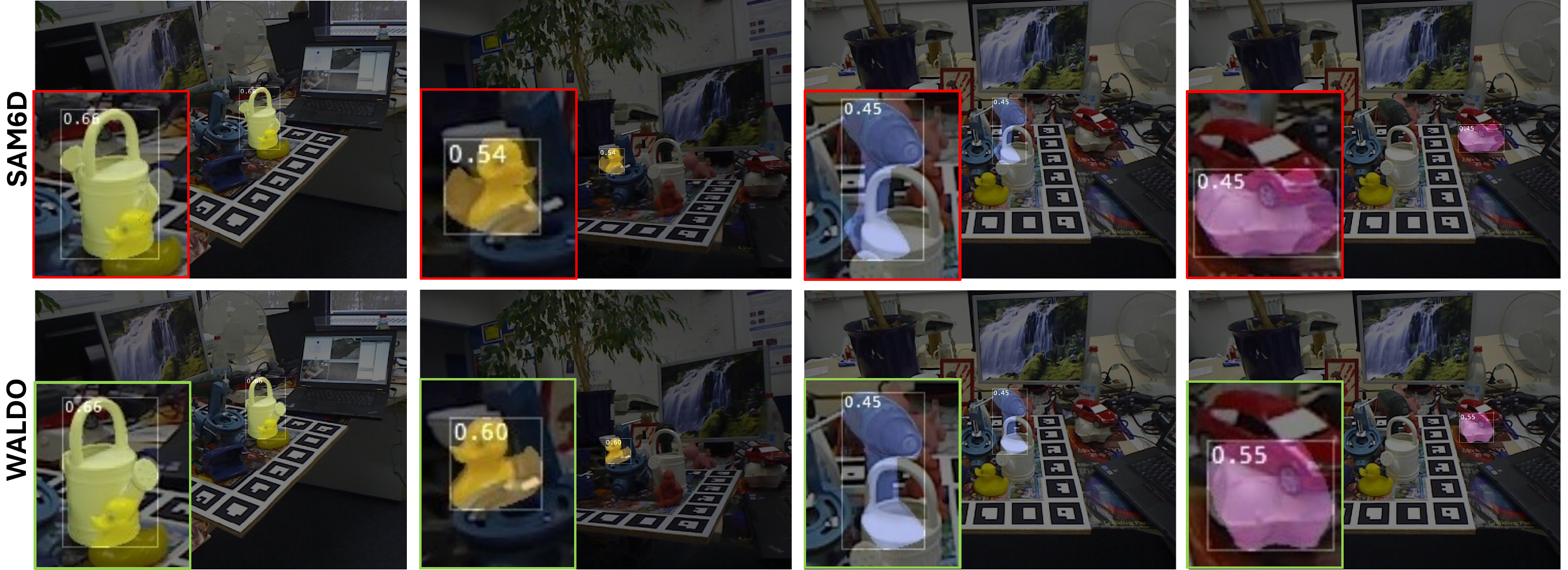}\vspace{-0.3cm}
    \caption{Qualitative comparison on LMO under occlusion. SAM6D  and WALDO. In these scenes, WALDO produces more reliable rotations.}
    \label{fig:lmo_qualitative}
    \vspace{-0.3cm}
\end{figure*}

\begin{figure}
    \centering
    \includegraphics[width=1\linewidth]{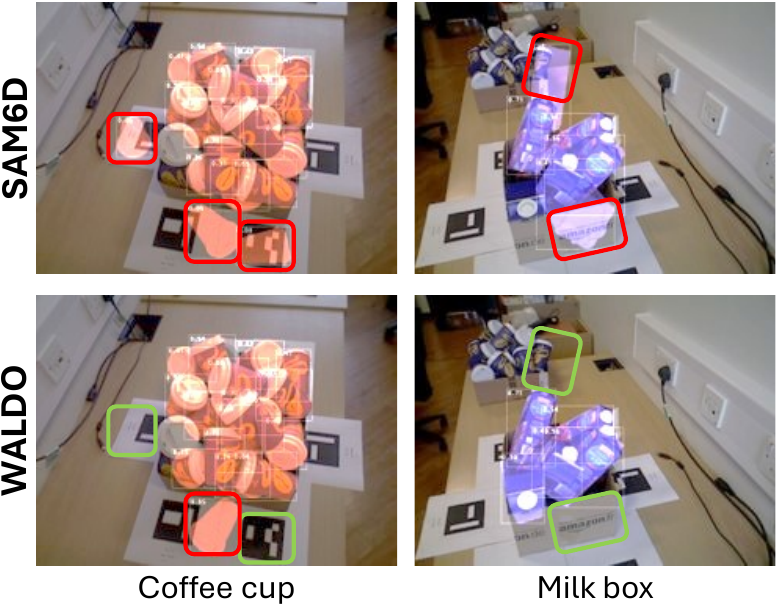}\vspace{-0.2cm}
    \caption{Qualitative comparison of WALDO vs. SAM6D on ICBIN. WALDO consistently produces fewer false positives in these scenes.}
    \label{fig:icbin_qualitative} \vspace{-0.4cm}
\end{figure}

We compare 6D pose estimation performance on BOP-Core* against state-of-the-art models (FRTPose, FoundationPose, FreeZe, and Co-op) and, in particular, our base model SAM6D (see Table~\ref{tab:uar_ar_s_results}). According to UAR, WALDO improves over SAM6D  on BOP-Core* by \(2\%\) while reducing per-image inference time from \(4.37\,\text{s}\) to \(1.53\,\text{s}\) (\(-65\%\)). The accuracy gain is larger on datasets with heavier occlusion, such as ICBIN where $+6\%$ improvement is achieved. We also observe $+4\%$ on YCBV and $+1\%$ on both LMO and T-LESS, while performance on TUDL remains on par. Further qualitative comparisons are given in Figure \ref{fig:lmo_qualitative} and \ref{fig:icbin_qualitative}.

Across BOP-Core* datasets, AR is consistently larger than UAR, indicating that standard AR overweights highly visible instances. For BOP-Core* with WALDO, AR exceeds UAR by \(0.172\) (\(+30.5\%\) relative to UAR). The gap is especially large on less-occluded sets, such as YCBV (\(+55.0\%\)) and TLESS (\(+62.1\%\)), and smaller on the highly occluded ICBIN (\(+13.2\%\)). Overall, the lower UAR indicates that our metric mitigates the visibility bias of standard AR, avoiding over-reporting and providing a fair assessment of performance.

Although the highest UAR on BOP-Core* is achieved by FRTPose (\(0.70\)), WALDO offers a favorable speed/accuracy trade-off, with \(1.53\,\text{s}\) runtime per image (a \(29.7\times\) speedup, i.e., \(-96.6\%\) latency) relative to FRTPose with a \(45.37\,\text{s}\) runtime per-image.

\begin{table*}[htbp]
\centering
\caption{Comparison of 6D pose estimatoion models on BOP-Core* and per-dataset using the conventional average recall (AR) and the proposed unbiased variant (UAR). Entries are reported as \emph{UAR/AR}. \emph{Time} denotes per-image inference runtime in second(s).}

\begin{tabular}{l |c c c c c |c c}
\toprule
Model & LMO & TLESS & TUDL & ICBIN & YCBV & BOP-Core*  & Time\\
\midrule
FRTPose & 0.68/0.79 & 0.61/0.76 & 0.91/0.93 & 0.67/0.71 & 0.61/0.90 & 0.70/0.82& 45.37\\
FoundationPose  & 0.58/0.75 & 0.45/0.64 & 0.83/0.92 & 0.42/0.50 & 0.59/0.88 & 0.57/0.74&29.32\\
FreeZe & 0.61/0.77 & 0.54/0.75 & 0.93/0.97 & 0.64/0.69 & 0.60/0.91 & 0.66/0.82& 24.89\\
Co-op & 0.55/0.73 & 0.45/0.69 & 0.92/0.92 & 0.52/0.63 & 0.60/0.89 & 0.61/0.77& 6.92\\
\midrule
SAM6D & 0.49/0.69 & 0.31/0.51 & \textbf{0.85}/0.90 & 0.50/0.58 & 0.52/0.84 & 0.54/0.71 & 4.37\\
\textbf{WALDO} & \textbf{0.50}/\textbf{0.71} & \textbf{0.32}/\textbf{0.53} & 0.85/\textbf{0.92} & \textbf{0.56}/\textbf{0.63}& \textbf{0.56}/\textbf{0.87} & \textbf{0.56}/\textbf{0.73}& \textbf{1.53}\\
\bottomrule
\end{tabular}\vspace{-0.4cm}
\label{tab:uar_ar_s_results}
\end{table*}

\subsubsection{Robustness to viewpoints}\label{sec:view}
\begin{figure}[t]
    \centering
    \includegraphics[width=0.8\linewidth]{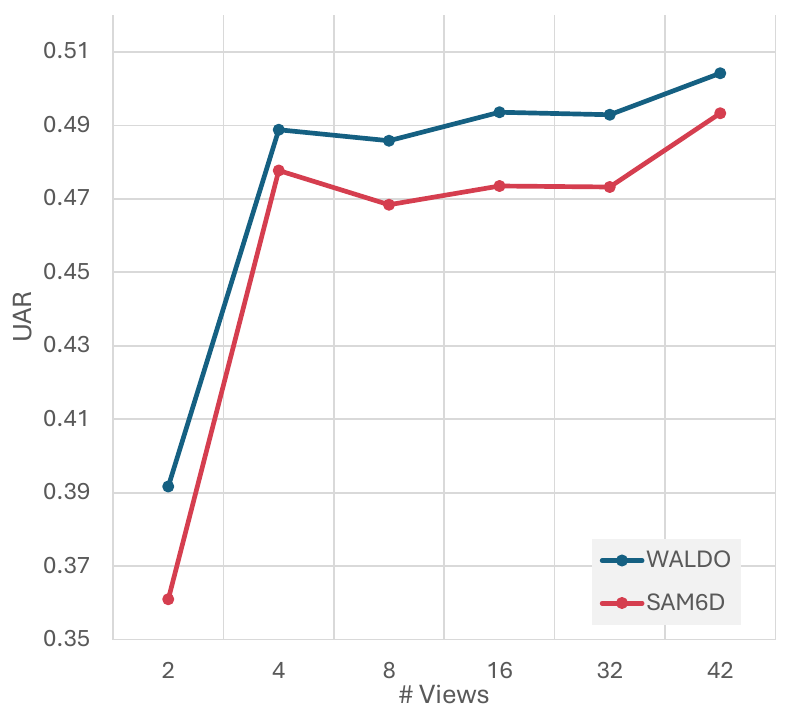}
    \caption{Comparison of the sensitivity of WALDO vs. SAM6D to the number of viewpoints used at inference to create object's point cloud.}
    \label{fig:uar_vs_views}
\end{figure}

We assess the sensitivity of estimated pose to the number of template viewpoints used at inference, varying from 2 to 42. We use at least 2 opposing views  to create a complete point cloud from the object templates. As shown in Figure~\ref{fig:uar_vs_views}, WALDO outperforms SAM6D across all view counts and follows a smooth increasing trend as more views are added, indicating lower sensitivity to viewpoints. By contrast, SAM6D shows a sharp drop after three views and fluctuates thereafter. These results suggest that WALDO maintains stable pose estimates with few views and continues to benefit from additional viewpoints.

\subsection{Ablations}\label{sec:ablation}
\noindent\textbf{Detector choice.}
 We ablate the effect of detection and segmentation accuracy on 6D pose recall in Table~\ref{tab:abl_det}. MUSE yields higher recall (UAR $=0.56$), outperforming NIDS ($0.54$) and Fast-SAM ($0.55$). Using ground-truth detections raises UAR to $0.70$, indicating remaining room for improving detection and segmentation. Although NIDS is slightly faster, MUSE provides better recall thus providing a better trade-off between accuracy and speed.

\begin{table}[htbp]
\centering
\caption{Effect of detector choice on 6D pose recall. We report UAR on BOP-Core* and per-image inference runtime in second(s).}

\begin{tabular}{lcc}
\toprule
Detector            & UAR & Time \\
\midrule
NIDS          & 0.54 &  \textbf{1.26}     \\
Fast-SAM          & 0.55  & 4.37        \\
MUSE     & \textbf{0.56}   & 1.53      \\
Ground Truth       & 0.70  & --        \\
\bottomrule
\end{tabular}\vspace{-0.3cm}
\label{tab:abl_det}
\end{table}

\noindent\textbf{Occlusion-focused augmentation.}
We ablate the effect of the proposed occlusion-aware augmentations in Table~\ref{tab:abl_aug}. We begin with a simple baseline where we directly use the features extracted from ViT and solve the \emph{3D-3D }correspondence with RANSAC (UAR $=0.33$). Then we add the course and dense point matching components and solve correspondence with SVD (UAR $=0.54$). Lastly we add the proposed occlusion-focused augmentations during training with negligible overhead (UAR $=0.55$). The augmentations further improve our strong Point Maching + SVD baseline, directly targeting occluded cases and doing so with no additional inference cost.

\begin{table}[htbp]
\centering
\caption{Effect of the proposed occlusion-focused augmentation on 6D pose recall reported as UAR on BOP-Core*.}

\begin{tabular}{lc}
\toprule
Method            & UAR \\
\midrule
VIT \& RANSAC      & 0.33         \\
+ Point Matching \& SVD   & 0.54         \\
+ Augmentation     & \textbf{0.55}         \\
\bottomrule
\end{tabular}\vspace{-0.3cm}
\label{tab:abl_aug}
\end{table}

\noindent\textbf{Dynamic dense sampling.}
We ablate the effect of replacing static uniform sampling with dynamic non-uniform sampling at the dense point matching stage in Table. \ref{tab:abl_DDS}. This replacement improves UAR on LMO, increasing from \(0.49\) to \(0.50\), indicating focusing the point matching process on the non-occluded parts of the objects results in higher recall. 

\begin{table}[htbp]
\centering
\caption{Comparison of dynamic non-uniform dense sampling (DNDS) and static uniform sampling. We report UAR on LMO for both methods.}
\begin{tabular}{lc}
\toprule
Method  & UAR \\
\midrule
Uniform Sampling & 0.49\\
DNDS & \textbf{0.50}\\
\bottomrule
\end{tabular}
\label{tab:abl_DDS}
\end{table}

\noindent\textbf{Multi-hypothesis at the coarse stage.}
We ablate the effect of the number of   initial course pose hypotheses $K$ on the final 6D pose recall in Table~\ref{tab:abl_k}. Increasing $K$ shows a gradual improvement up to $K{=}8$ (UAR $=0.5045$), followed by a mild decline as $K$ increases further. This suggests that considering multiple initial hypothesis and selecting the final pose after refinement stage, allows the pipeline to recover from early stage errors, but only if $k$ is not increased excessively.

\begin{table}[htbp]
\centering
\caption{Ablation on LMO showing 6D pose recall (UAR) as a function of the number of coarse-stage hypotheses (\(K\)).}
\resizebox{\linewidth}{!}{%
\begin{tabular}{l|cccccc}
\toprule
$K$  & 1& 2& 4& 8& 16& 32 \\
\midrule
UAR & 0.4933 & 0.4901 & 0.4959 & \textbf{0.5045} & 0.5022 & 0.4989\\
\bottomrule
\end{tabular}%
}
\label{tab:abl_k}
\end{table}

\noindent\textbf{Iterative refinement depth.}
We ablate the effect of iteratively refining the final pose by reusing it as an initial hypothesis in Table \ref{tab:abl_n}. Increasing the number of refinement steps $N$ initially reduces accuracy at $N{=}2$–$4$, then recovers and surpasses the single-step baseline, peaking at $N{=}32$ (UAR $=0.5015$). This is because early refinements move the pose away from a suboptimal solution before converging toward a better one.

\begin{table}[htbp]
\centering
\caption{Ablation on LMO showing 6D pose recall (UAR) as a function of the number of refinement steps (\(N\)).}\vspace{-0.1cm}
\resizebox{\linewidth}{!}{%
\begin{tabular}{l|cccccc}
\toprule
$N$  & 1& 2& 4& 8& 16 & 32\\
\midrule
UAR & 0.4933 & 0.4810 & 0.4853 & 0.4956 & 0.4957 & \textbf{0.5015}\\
\bottomrule
\end{tabular}
}\vspace{-0.3cm}
\label{tab:abl_n}
\end{table}

\section{Conclusion}
We proposed WALDO, a model-based 6D pose estimation framework that makes the standard detect–segment–hypothesize–refine pipeline more robust for occlusion handling. Our method, first estimates occlusion probabilities and uses a dynamic non-uniform dense sampling strategy to focus computation on visible regions. Next, to avoid brittle decisions, it maintains multiple pose candidates, rank them by confidence, and refines them iteratively, which preserves accuracy even when parts of the object are hidden. In addition, we proposed training-time augmentations that realistically simulate noise and occlusion. For evaluation, we introduced an occlusion-aware evaluation metric that reduces the optimistic bias of protocols dominated by highly visible instances, giving a fairer view of progress under occlusion. By evaluating across benchmarks, WALDO improves accuracy by more than \textbf{5\%} on ICBIN and more than \textbf{2\%} on BOP, while achieving approximately \(\mathbf{3\times}\) faster inference. These gains translate to more reliable poses in cluttered, real-world scenes.

{
    \small
    \bibliographystyle{ieeenat_fullname}
    \bibliography{main}
}

\end{document}


\maketitle

\clearpage
\setcounter{page}{1}
\maketitlesupplementary

\section{Rationale}
\label{sec:rationale}
Having the supplementary compiled together with the main paper means that:
\begin{itemize}
\item The supplementary can back-reference sections of the main paper, for example, we can refer to \cref{sec:intro};
\item The main paper can forward reference sub-sections within the supplementary explicitly (e.g. referring to a particular experiment); 
\item When submitted to arXiv, the supplementary will already included at the end of the paper.
\end{itemize}
To split the supplementary pages from the main paper, you can use \href{https://support.apple.com/en-ca/guide/preview/prvw11793/mac#:~:text=Delete%20a%20page%20from%20a,or%20choose%20Edit%20%3E%20Delete).}{Preview (on macOS)}, \href{https://www.adobe.com/acrobat/how-to/delete-pages-from-pdf.html#:~:text=Choose%20%E2%80%9CTools%E2%80%9D%20%3E%20%E2%80%9COrganize,or%20pages%20from%20the%20file.}{Adobe Acrobat} (on all OSs), as well as \href{https://superuser.com/questions/517986/is-it-possible-to-delete-some-pages-of-a-pdf-document}{command line tools}.

\begin{table*}
    \centering
    \resizebox{0.97\textwidth}{!}{
    \begin{tabular}{l|c|ccccccccccc|c}
      \toprule
      Dataset & Model & AP & APD1 & APD2 & APD3 & APD4 & APD5 & APD6 & APD7 & APD8 & APD9 & APD10 & mAPD\\
      \midrule
      \multirow{2}{*}{BOP-Core*} 
      & SAM6D      & 0.491 & -1.000 & 0.000 & 0.003 & 0.019 & 0.099 & 0.135 & 0.160 & 0.246 & 0.366 & 0.567 & 0.202\\
       & MUSE     & 0.508 & -1.000 & 0.001 & 0.002 & 0.058 & 0.109 & 0.146 & 0.198 & 0.248 & 0.386 & 0.599 & 0.218\\
       & NIDS     & 0.463 & -1.000 & 0.001 & 0.003 & 0.039 & 0.108 & 0.159 & 0.187 & 0.266 & 0.373 & 0.574 & 0.214\\
       & Fast-SAM & 0.456 & -1.000 & 0.000 & 0.001 & 0.049 & 0.102 & 0.134 & 0.216 & 0.243 & 0.361 & 0.539 & 0.216\\
       & CNOS     & 0.425 & -1.000 & 0.000 & 0.001 & 0.071 & 0.061 & 0.103 & 0.147 & 0.170 & 0.288 & 0.506 & 0.169\\
      \bottomrule
    \end{tabular}}
    \caption{Average detection AP performance over all datasets in BOP-Core*. For each metric, undefined values (``-1'') are omitted from the averaging.}
    \label{tab:detection_ap_avg_defined}
\end{table*}

\begin{table*}
    \centering
    \resizebox{0.97\textwidth}{!}{
    \begin{tabular}{l|c|ccccccccccc|c}
      \toprule
      Dataset & Model & AR & ARD1 & ARD2 & ARD3 & ARD4 & ARD5 & ARD6 & ARD7 & ARD8 & ARD9 & ARD10 & mARD\\
      \midrule
      \multirow{2}{*}{BOP-Core*} 
      & SAM6D      & 0.576 & -1.000 & 0.025 & 0.085 & 0.265 & 0.328 & 0.398 & 0.553 & 0.529 & 0.614 & 0.679 & 0.433\\
       & MUSE     & 0.594 & -1.000 & 0.022 & 0.066 & 0.266 & 0.344 & 0.401 & 0.568 & 0.566 & 0.632 & 0.705 & 0.444\\
       & NIDS     & 0.555 & -1.000 & 0.034 & 0.060 & 0.274 & 0.311 & 0.381 & 0.534 & 0.525 & 0.594 & 0.679 & 0.421\\
       
       & Fast-SAM & 0.528 & -1.000 & 0.018 & 0.031 & 0.224 & 0.261 & 0.343 & 0.502 & 0.474 & 0.562 & 0.636 & 0.384\\
       & CNOS     & 0.519 & -1.000 & 0.017 & 0.031 & 0.225 & 0.261 & 0.338 & 0.492 & 0.450 & 0.542 & 0.626 & 0.374\\
      \bottomrule
    \end{tabular}}
    \caption{Average detection AR performance over all datasets in BOP-Core*. For each metric, undefined values (``-1'') are omitted in the averaging.}
    \label{tab:detection_ar_avg}
\end{table*}

\begin{table*}
    \centering
    \resizebox{0.97\textwidth}{!}{
    \begin{tabular}{l|c|ccccccccccc|c}
      \toprule
      Dataset & Model & AP & APD1 & APD2 & APD3 & APD4 & APD5 & APD6 & APD7 & APD8 & APD9 & APD10 & mAPD\\
      \midrule
      \multirow{5}{*}{LMO} & MUSE & 0.478 & -1.000 & 0.002& 0.001& 0.006& 0.033& 0.113& 0.204& 0.358 &0.425& 0.611& 0.195\\
       & NIDS & 0.457 & -1.000 & 0.003 & 0.007 & 0.027 & 0.056 & 0.125 & 0.209 & 0.330 & 0.389 & 0.553 & 0.189\\
       & SAM  & 0.468 &  -1.000	&0.001 & 0.005 & 0.008 & 0.051& 0.125 & 0.212& 0.316 & 0.390 & 0.576 & 0.187\\
       & Fast-SAM & 0.430 & -1.000 & 0.001	&0.002	&0.004	&0.037	&0.090	&0.172	&0.295	&0.359	&0.534	&0.166\\
       & CNOS & 0.404 & -1.000	&0.001	&0.002	&0.006	&0.028	&0.074	&0.146	&0.247	&0.312	&0.507	&0.147\\
        \midrule
      \multirow{5}{*}{TLESS} & MUSE & 0.451 & -1.000	&0.000	&0.002	&0.014	&0.026	&0.042	&0.145	&0.191	&0.301	&0.526	&0.139 \\
       & NIDS & 0.493 & -1.000	&0.001	&0.001	&0.012	&0.046	&0.077	&0.201	&0.254	&0.360	&0.571	&0.169\\
       & SAM  & 0.451 & -1.000	&0.000	&0.001	&0.007	&0.021	&0.030	&0.134	&0.217	&0.320	&0.523	&0.139\\
       & Fast-SAM & 0.420 & -1.000	&0.000	&0.001	&0.018	&0.029	&0.045	&0.136	&0.180	&0.313	&0.492	&0.135\\
       & CNOS & 0.374 & -1.000	&0.000	&0.001	&0.015	&0.022	&0.039	&0.105	&0.129	 &0.229	&0.431	&0.108\\
        \midrule
      \multirow{5}{*}{TUDL} & MUSE & 0.565 & -1.000	&-1.000	&-1.000	&-1.000	&-1.000	&-1.000	&0.100	&0.094	&0.328	&0.574	&0.274\\
       & NIDS & 0.486 & -1.000&-1.000&-1.000	&-1.000	&-1.000	&-1.000	& 0.036	&0.163	&0.355	&0.563	&0.279\\
       & SAM  & 0.569 &-1.000	&-1.000	&-1.000	&-1.000	&-1.000	&-1.000	&0.003	&0.167	&0.346	&0.577	&0.273\\
       & Fast-SAM & 0.517 & -1.000	&-1.000	&-1.000	&-1.000	&-1.000	&-1.000	&0.325	&0.219	&0.350	&0.525	&0.355\\
       & CNOS & 0.480 & -1.000	&-1.000	&-1.000	&-1.000	&-1.000	&-1.000	&0.132	&0.034	&0.247	&0.486	&0.225\\
        \midrule
      \multirow{5}{*}{ICBIN} & MUSE & 0.375 & -1.000	&0.000	&0.004	&0.007	&0.043	&0.083	&0.120	&0.147	&0.288	&0.547	&0.138 \\
       & NIDS & 0.257 & -1.000	&0.000	&0.005	&0.008	&0.026	&0.082	&0.100	&0.141	&0.235	&0.466	&0.118\\
       & SAM  & 0.358 & -1.000	&0.000	&0.004	&0.007	&0.053	&0.118	&0.154	&0.178	&0.314	&0.498	&0.147\\
       & Fast-SAM & 0.293 &  -1.000	&0.000	&0.002	&0.004	 &0.019	&0.070	&0.101	&0.148	&0.264	&0.464	&0.119\\
       & CNOS & 0.270 & -1.000	&0.000	&0.002	&0.003	&0.013	&0.055	&0.069	 &0.112	&0.233	&0.419	&0.101\\
        \midrule
      \multirow{5}{*}{YCBV} & MUSE & 0.672 & -1.000	&0.000	&0.000	&0.203	&0.334	&0.345	&0.422	&0.449	&0.588	&0.736	&0.342 \\
       & NIDS & 0.621 & -1.000	&0.001	&0.000	&0.107	&0.302	&0.352	&0.391	&0.442	&0.526	&0.716	&0.315\\
       & SAM  & 0.607 & -1.000	&0.000	&0.000	&0.055	&0.270	&0.265	&0.296	&0.353	&0.462	&0.662	&0.262\\
       & Fast-SAM & 0.621 & -1.000	&0.000	&0.000	&0.169	&0.324	&0.330	&0.344	&0.373	&0.517	&0.681	&0.304\\
       & CNOS & 0.599 & -1.000	&0.000	&0.000	&0.259	&0.180	&0.242	&0.283	&0.329	&0.418	&0.685	&0.266\\
      
      \bottomrule
    \end{tabular}}
    \vspace{-0.2cm}
    \caption{Instance segmentation results of different methods on the seven core datasets of the BOP benchmark \cite{BOP}. We report the mean Average Precision (mAP) scores at different Intersection-over-Union (IoU) values ranging from 0.50 to 0.95 with a step size of 0.05. }
    \vspace{-0.25cm}
    \label{tab:detection_ap}
\end{table*}

\begin{table*}
    \centering
    \resizebox{0.97\textwidth}{!}{
    \begin{tabular}{l|c|ccccccccccc|c}
      \toprule
      Dataset & Model & AR & ARD1 & ARD2 & ARD3 & ARD4 & ARD5 & ARD6 & ARD7 & ARD8 & ARD9 & ARD10 & mARD\\
      \midrule
      \multirow{5}{*}{LMO} & MUSE & 0.538 & -1.000	&0.029	&0.072	&0.175	&0.213	&0.376	&0.407	&0.526	&0.584	&0.711	&0.344\\
       & NIDS & 0.519 & -1.000	&0.027	 &0.050	&0.195	&0.236	&0.312	&0.353	&0.475	&0.524	&0.650	&0.314\\
       & SAM  & 0.544 &  -1.000	&0.042	&0.134	&0.209	&0.263	&0.367	&0.400	&0.523	&0.575	&0.698	&0.357\\
       & Fast-SAM & 0.501 & -1.000	&0.027	&0.030	&0.105	&0.182	&0.306	&0.335	&0.473	&0.537	&0.658	&0.295\\
       & CNOS & 0.497 & -1.000	&0.027	&0.023	&0.119	&0.178	&0.304	&0.325	&0.467	&0.525	&0.658	&0.292\\
        \midrule
      \multirow{5}{*}{TLESS} & MUSE & 0.564 & -1.000	&0.031	&0.042	&0.159	&0.291	&0.307	&0.465	&0.537	&0.605	&0.649	&0.343 \\
       & NIDS & 0.604 & -1.000	&0.047	&0.039	&0.212	&0.267	&0.367	&0.501	&0.561	&0.618	&0.689	&0.367\\
       & SAM  & 0.553 & -1.000	&0.031	&0.054	&0.176	&0.239	&0.304	&0.507	&0.536	&0.569	&0.630	&0.338\\
       & Fast-SAM & 0.511 & -1.000	&0.022	&0.032	&0.120	&0.186	&0.300	&0.429	&0.460	&0.553	&0.590	&0.299\\
       & CNOS & 0.499 & -1.000	&0.019	&0.032	&0.123	&0.186	&0.289	&0.421	&0.438	&0.539	&0.577	&0.292\\
        \midrule
      \multirow{5}{*}{TUDL} & MUSE & 0.652 & -1.000	&-1.000	&-1.000	&-1.00	&-1.000	&-1.000	&0.800	&0.518	&0.583	&0.665	&0.642\\
       & NIDS & 0.598 & -1.000	&-1.000	&-1.000	&-1.00	&-1.000	&-1.000	&0.800	&0.461	&0.557	&0.625	&0.611\\
       & SAM  & 0.650 &-1.000	&-1.000	&-1.000	&-1.00	&-1.000	&-1.000	&0.800	&0.486	&0.605	&0.660	&0.638\\
       & Fast-SAM &  0.577 & -1.000	&-1.000	&-1.000	&-1.00	&-1.000	&-1.000	&0.800	&0.421	&0.509	&0.590	&0.580\\
       & CNOS & 0.566 & -1.000	&-1.000	&-1.000	&-1.00	&-1.000	&-1.000	&0.800	&0.364	&0.496	&0.580	&0.560\\
        \midrule
      \multirow{5}{*}{ICBIN} & MUSE & 0.488 & -1.000	&0.027	&0.150	&0.209	&0.386	&0.414	&0.527	&0.596	&0.626	&0.719	&0.406 \\
       & NIDS & 0.361 & -1.000	&0.023	&0.150	&0.171	&0.257	 &0.351	&0.445	&0.541	&0.553	&0.661	&0.350\\
       & SAM  & 0.454 & -1.000	&0.026	&0.152	&0.148	&0.327	&0.397	&0.501	&0.520	&0.597	&0.682	&0.372\\
       & Fast-SAM & 0.363 & -1.000	&0.023	&0.063	&0.113	&0.192	&0.273	&0.368	&0.430	&0.503	&0.602	&0.285\\
       & CNOS & 0.355 & -1.000	&0.023	&0.068	&0.103	&0.196	&0.267	&0.360	&0.413	&0.503	&0.577	&0.279\\
        \midrule
      \multirow{5}{*}{YCBV} & MUSE & 0.730 & -1.000	&0.000	&0.000	&0.521	&0.486	&0.505	&0.643	&0.655	&0.763	&0.780	&0.484 \\
       & NIDS & 0.694 &-1.000	&0.039	&0.000	&0.518	&0.482	&0.493	&0.572	&0.589	&0.720	&0.771	&0.465\\
       & SAM  & 0.681 & -1.000	&0.000	&0.000	&0.525	&0.482	&0.522	&0.557	&0.580	&0.725	&0.727	&0.458\\
       & Fast-SAM & 0.687 & -1.000	&0.000	&0.000	&0.556	&0.485	&0.493	&0.578	&0.586	&0.710	 &0.740	&0.461\\
       & CNOS & 0.677 & -1.000	&0.000	&0.000	&0.556	&0.485	&0.493	&0.554	&0.568	&0.648	&0.740	&0.449\\
      
      \bottomrule
    \end{tabular}}
    \vspace{-0.2cm}
    \caption{ }
    \vspace{-0.25cm}
    \label{tab:detection_ar}
\end{table*}

\begin{table*}[htbp]
\centering
\begin{tabular}{l l c c c c c c c}
\toprule
Model & Metric &BOP-Core & LMO & TLESS & TUDL & ICBIN & YCBV  & Time\\
\midrule
\multirow{2}{*}{FRTPose} 
  & UAR &0.7008& 0.6831 & 0.6188 & 0.9133 & 0.6753 & 0.6133 & 45.37\\
  & AR          &0.8260& 0.7950 & 0.7676 & 0.9396 & 0.7190 & 0.9089 & \\
\midrule
\multirow{2}{*}{Foundation Pose} 
  & UAR &0.5783& 0.5889 & 0.4528 & 0.8312 & 0.4209 & 0.5976 & 29.32\\
  & AR          &0.7444 & 0.7564 & 0.6459 & 0.9230 & 0.5078 & 0.8891 & \\
\midrule
\multirow{2}{*}{Freeze} 
  & UAR &0.6693& 0.6101 & 0.5497 & 0.9354 & 0.6450 & 0.6064 & 24.89\\
  & AR          &0.8227& 0.7711 & 0.7548 & 0.9765 & 0.6968 & 0.9145 & \\
\midrule
\multirow{2}{*}{Co-op} 
  & UAR &0.6119& 0.5561 & 0.4594 & 0.9207 & 0.5228 & 0.6004 & 6.92\\
  & AR          &0.7787& 0.7376 & 0.6946 & 0.9288 & 0.6349 & 0.8976 & \\
\midrule
\multirow{2}{*}{VIT + RANSAC} 
  & UAR &0.3379 & 0.2929 & 0.2171 & 0.4033 & 0.3519 & 0.4242 & 4.83\\
  & AR          &0.5256 & 0.4802 & 0.3954 & 0.6025 & 0.4492 & 0.7009 & \\
\midrule
\multirow{2}{*}{SAM6D} 
  & UAR &0.5402 & 0.4933 & 0.3131 & 0.8592 & 0.5067 & 0.5286 & 4.37\\
  & AR          &0.7100& 0.6988 & 0.5151 & 0.9041 & 0.5876 & 0.8446 & \\
\midrule
\multirow{2}{*}{SAM6D-FT} 
  & UAR &0.5541& 0.5131 & 0.3196 & 0.8734 & 0.5115 & 0.5530 & 4.37\\
  & AR          &0.7187& 0.7162 & 0.5185 & 0.9130 & 0.5953 & 0.8504 & \\
  \midrule
\multirow{2}{*}{SAM6D-FT-MUSE} 
  & UAR &0.5632& 0.5042 & 0.3293 & 0.8580 & 0.5604 & 0.5643 & 1.528\\
  & AR          &0.7351& 0.7108 & 0.5338 & 0.9217 & 0.6346 & 0.8746 & \\
   \midrule
\multirow{2}{*}{SAM6D-FT-NIDS} 
  & UAR &0.5426& 0.4689 & 0.3385 & 0.8456 & 0.4944 & 0.5655 & 1.266\\
  & AR          &0.7072& 0.6588 & 0.5414 & 0.8872 & 0.5730 & 0.8756 & \\
     \midrule
\multirow{2}{*}{SAM6D-FT-GT} 
  & UAR &0.7073& 0.6513 & 0.6455 & 0.9393 & 0.6952 & 0.6053 & \\
  & AR          &0.8415& 0.8017 & 0.7875 & 0.9757 & 0.7357 & 0.9071 & \\
\bottomrule
\end{tabular}
\caption{Unbiased AR and AR on multiple datasets for multiple models.}
\label{tab:ar_results}
\end{table*}